# EMR-based medical knowledge representation and inference via Markov random fields and distributed representation learning

**Chao Zhao**                                                                                    zhaochaocs@gmail.com
**Jingchi Jiang**                                                                          jiangjingchi0118@163.com
**Yi Guan**[*]                                                                                      guanyi@hit.edu.cn
*School of Computer Science and Technology*
*Harbin Institute of Technology*
*Harbin, Heilongjiang, 150001, CHN*

## Abstract

**Objective:** Electronic medical records (EMRs) contain an amount of medical knowledge which can be used for clinical decision support (CDS). Our objective is a general system that can extract and represent these knowledge contained in EMRs to support three CDS tasks: test recommendation, initial diagnosis, and treatment plan recommendation, with the given condition of one patient.

**Methods:** We extracted four kinds of medical entities from records and constructed an EMR-based medical knowledge network (EMKN), in which nodes are entities and edges reflect their co-occurrence in a single record. Three bipartite subgraphs (bi-graphs) were extracted from the EMKN to support each task. One part of the bi-graph was the given condition (e.g., symptoms), and the other was the condition to be inferred (e.g., diseases). Each bi-graph was regarded as a Markov random field to support the inference. Three lazy energy functions and one parameter-based energy function were proposed, as well as two knowledge representation learning-based energy functions, which can provide a distributed representation of medical entities. Three measures were utilized for performance evaluation.

**Results:** On the initial diagnosis task, 80.11% of the test records identified at least one correct disease from top 10 candidates. Test and treatment recommendation results were 87.88% and 92.55%, respectively. These results altogether indicate that the proposed system outperformed the baseline methods. The distributed representation of medical entities does reflect similarity relationships in regards to knowledge level.

---

[*]. corresponding author

**Conclusion:** Combining EMKN and MRF is an effective approach for general medical knowledge representation and inference. Different tasks, however, require designing their energy functions individually.

# 1. Introduction

Clinical decision support systems (CDSS) aim to provide clinicians or patients with computer-generated clinical knowledge and patient-related information that can be intelligently filtered or presented at appropriate times, to enhance patient care[1]. A core component of CDSS is the knowledge base, which was once established and updated manually by clinical experts; but is trying to be generated and managed automatically nowadays. This often includes natural language processing (NLP) techniques for mining clinical knowledge to drive CDSS from medical free-text[2], such as medical literature and electronic medical records (EMRs). Focused on the former source, Text REtrieval Conference (TREC) CDS track expected to develop a retrieval-based system to solve three following problems by returning the most relevant biomedical articles [3]:

- Determining a patient's most likely diagnosis given a list of symptoms

- Deciding on the most effective treatment plan for a patient with a known condition

- Determining if a particular test is indicated for a given situation

We would assert, however, that it is possible to shrink the information granularity from articles to medical entities. According to the patient condition, the CDSS can directly provide the proper investigations, diagnosis results, and ordered treatment plans, rather than simply the relevant literatures from which the clinician must draw the necessary information. The EMR is a credible source of medical knowledge for this purpose - it is the storage of all of a given patient's health care data and medical history of a patient in an electronic format. These data include abundant medical entities, such as the current clinical diagnosis, medical history, results of investigations, treatment plans, and so on[4, 5]. These entities, and relationships between entities, are the primary carriers of medical knowledge in EMR, and can be extracted by the information extraction technique[6, 7, 8]. It shows the possibility of acquiring and organizing medical knowledge automatically based on the EMR. After the information is extracted, the two subsequent key problems are (1) representing medical knowledge via these entities and entity relationships, and (2) making medical inferences according to this representation.

Several machine learning based solutions have been proposed to the above two problems[9], such as the statistical classifiers, association rules, Bayesian networks and so on. Advancements in representation learning and deep learning on NLP have also provided a new approach to CDS. Most methods focus only on one specific disease, however, due to limitations inherent to the model itself or the computational complexity, and universal support systems for general practice remain elusive. We believe it prudent to construct such a general system for two main reasons. First, this kind of system can respond to the demands of ordinary people suffering any problematic symptom. They may want to research independently before seeing a doctor. Second, the general system can yield initial results that may better support real applications than specialized CDS systems. The latter require much more patient information beyond just his or her symptoms and test results to guarantee precise results.

In this paper, we make a preliminary attempt to represent medical knowledge from EMR, to resolve the three problems proposed at the beginning of the paper in a medical entity level. We first represent the medical knowledge using an EMR-based medical knowledge network (EMKN), and then regard it as a Markov random field (MRF) for inference tasks. In the EMKN, nodes are medical entities and edges are entity co-occurrence relationships. The MRF describes the probability distribution among these entities and makes probabilistic inferences according to the pre-defined energy functions.

Our main contributions are three-fold:

- We proposed a universal EMR-based clinical decision support method using EMKN and MRF. This method takes only the corresponding medical entities as inputs and is not restricted to certain diseases.

- We derived a learning algorithm for arbitrarily derivable energy functions, and integrated the knowledge representation learning approaches into the MRF, to obtain a distributed representation of medical entities.

- We applied the inference architecture to three CDS tasks: test suggestion, initial diagnosis and treatment plan suggestion. This allow us to experimentally demonstrate the efficiency of the proposed method on actual clinical records.

The remainder of this paper is structured as follows. In Section 2, we give a brief review of existing CDS systems as well as the related works of representation learning. In Section 3, we describe the details of construction of EMKN, as well as the inference and learning algorithms based on MRF. Section 4 introduces two distributed representation methods of medical entities to MRF. We evaluated those methods using actual records, as described in Section 5; the results are discussed at length in Section 6. A brief conclusion and discussion on future research directions are presented in Section 7.

## 2. Related Works

The three problems referred to in Section 1 can be further generalized as one problem: to provide the best possible clinical recommendations (medical investigations, possible diagnosis, and treatment plans) for a given patient's condition. This section introduces previous works relevant to medical knowledge representation and decision making in regards to this problem.

Many of the existing CDS systems focus on one (or one kind of) disease, and adopt classification strategies to solve this problem. Some typical patient features (e.g. signs, symptoms, test results) are extracted with the help of domain expert knowledge, and then transformed and selected. After the feature engineering, the disease condition can be determined by general classifiers like the logistic regression[10, 11], neural network[12, 13] or naïve Bayes classifier[14]. For example, [10] constructed a series of classifiers to predict the estimated glomerular filtration rate (eGFR) of kidney transplant patients, with the help of 56 selected features from the donor and recipient.

Other researchers have attempted to develop models without requiring the manual input of prior knowledge and to depict the relationships among clinical events directly from the data. Association rules mining is a typical approach to identifying relationships of clinical events pairs [15, 16, 17]. Bayesian networks (or "probabilistic graphical models", more generally) can also be used to represent the relationships of medical events[18, 19, 20]. For example, [20] utilized a Bayesian network to implement an adaptive recommendation system to recommend a next order of treatment menu, based on the previous orders. Compared to association rules mining, Bayesian networks can properly account for transitive associations and co-varying relationships among variables.

In attempting to diagnose more than one disease via the approaches described above, the size of the feature set and the number of random variables would be excessive, and neither binary classifiers or Bayesian networks is a good choice. The former would suffer from curse of dimensionality and class imbalance, and the latter are limited by the computational complexity of inference and learning[21]. Non-classification based models are more appropriate. For example, [22] developed a non-disease-specific AI simulation framework via Markov decision process to evaluate the consequences of specific treatment plans; [23] analyzed clinical pathways from clinical workflow log using process mining approaches.

Recent development in representation learning and deep learning have opened new opportunities for medical knowledge representation. Representation learning aims to learn a good representation of the data, which can make it easier to extract useful information when building classifiers[24]. One widely used representation technique is deep learning[25, 26], which has been particularly successful in a variety of artificial intelligence (AI) fields[27, 28], including NLP[29, 30, 31], where researchers attempted to map the word $w$ to a low-dimensional, dense vector $\boldsymbol{w} \in \mathbb{R}^n$. Different entries of the vector depict the word's features from various aspects. This so-called *distributed representation* method can mitigate data sparsity and improve the generalization power of the model to which it is applied. CBOW and skip-gram[31, 32] are two popular algorithms to obtain such representations.

In medical text processing, researchers have attempted to learn the distributed representation of medical terms using similar approaches. Several have fed unstructured medical copora directly to word2vec toolkits[33], but it is more common to extract the medical concepts from raw text first, and then to learn the representation over the temporal medical concept sequences[34, 35]. The obtained medical concept embeddings can be further applied to the relation extraction[36], patient intention detection[37], and even diagnosis and risk prediction[38, 35, 39]. [35], for example, modeled temporal relations among medical events using recurrent neural networks to efficiently detect heart failure onset. Though the final layer of their model was still a classifier, it only required the clinical events as inputs and all the features used for classification were learned automatically.

Representation learning can also be applied to knowledge representation[40, 41, 42, 43]. Part of human knowledge can be represented in the form of a relation triple $(e_h, r, e_t)$, where there is a certain relationship $r$ from the head entity $e_h$ to the tail entity $e_t$. Knowledge representation learning (KRL) is deployed to obtain low-dimensional embeddings for entities and relationships of these triples. Typical KRL models include latent factor models (LFMs)[42] and translating embedding (TransE) models [43]. TransE is especially popular due to its prediction accuracy and computational efficiency. To the best of our knowledge, however, these methods have not been applied to learning medical concept embeddings.

## 3. Methods

The network is a convenient tool for modeling and visualizing entities with complex relationships. In this work, we began by organizing a series of medical entities into a network.

### 3.1 EMR-based Medical Knowledge Network

We proposed the EMKN, an EMR-based medical knowledge network for knowledge representation from EMR, in [44]. This section gives a brief review and supplementary information about this network.

The corpus we used contained 992 de-identified clinical records[45], which were retrieved from The Second Affiliated Hospital of Harbin Medical University. We manually extracted the medical entities and its modifiers[1]. Medical entities were roughly split into five categories: *symptom*, *test*, *test result*, *disease* and *treatment*. The modifiers included *present*, *possible*, *absent* and the other four modifiers. Based on these entities, we constructed EMKN, where nodes served as medical entities and edges were co-occurrence relationships among entities in one single record.

In this work, We extract three bigraphs from EMKN as listed in Table 1, to support the three problems named in section 1. We denote the bigraph by $G = (\boldsymbol{X}, \boldsymbol{Y})$, where $\boldsymbol{X}$ is the entity set we have observed and $\boldsymbol{Y}$ is our corresponding recommendation items. The diagnosis task, for example, depends only on the SD-EMKN. $\boldsymbol{X}$ denotes the symptom and test result entities, and $\boldsymbol{Y}$ denotes the disease entities. Figure 1 is a visualization of the ENK with its three bigraphs.

---
1. Although an information extraction system has been developed with the help of these annotated data and a larger database is available, we still used the manually annotated data to eliminate any interferences with the automatic results.

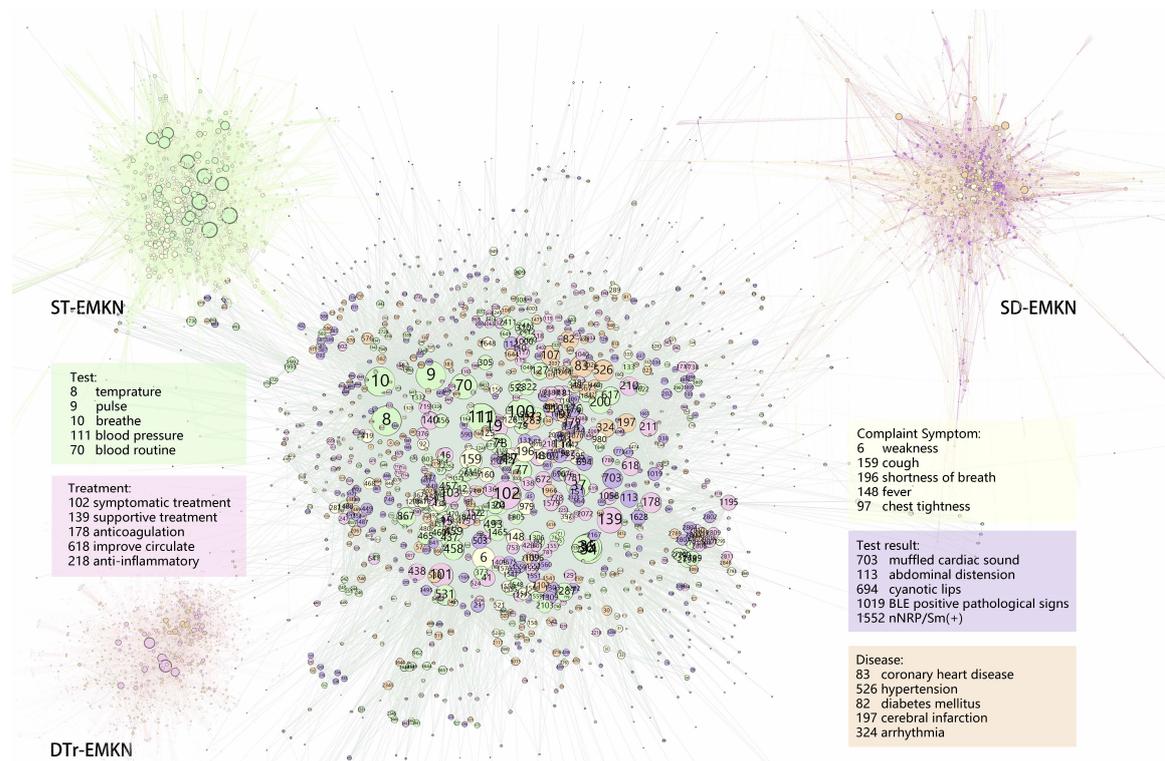

Figure 1: EMKN visualized in Gephi. Nodes sizes are proportional to their degree, and node type is indicated by color. Each node is labeled with a corresponding digital id. Corresponding medical entities are provided for several nodes.

Table 1: Statistical quantities of three EMKN bigraphs. We list the node size of each part, as well as the mean and median of the node degree.

| subgraph name | **X** | | | | **Y** | | | |
|---|---|---|---|---|---|---|---|---|
| | type | size | degree mean | degree median | type | size | degree mean | degree median |
| SD-EMKN | symptom test result | 2148 | 6.85 | 4 | disease | 1208 | 12.19 | 8 |
| DTr-EMKN | disease | 667 | 8.25 | 6 | treatment | 263 | 10.46 | 5 |
| ST-EMKN | symptom | 811 | 10.2 | 8 | test | 538 | 15.37 | 5 |

### 3.2 From EMKN to MRF

MRF defines the joint probability among variables $\boldsymbol{A} = (A_1, A_2, \cdots, A_n)$ in terms of an undirected graph. Formally, the joint probability of $\boldsymbol{A}$ can be written as

$$P(\boldsymbol{A}) = \frac{1}{Z} \prod_{C \in \mathcal{C}} \phi_C(\boldsymbol{A}_C), \tag{1}$$

where $\mathcal{C}$ is the set of maximal cliques in the graph and $\phi_C(\boldsymbol{A}_C)$ is the corresponding potential of clique $C$. The potential function $\phi_C : \boldsymbol{A}_C \to \mathbb{R}^+$ defines a map from clique to a positive real number. The larger the value of $\phi_C(\boldsymbol{A}_C = \boldsymbol{a}_C)$, the more likely that $\boldsymbol{A}_C = \boldsymbol{a}_C$. To ensure positivity of $\phi_C(\cdot)$, we rewrote $\phi_C(\cdot) = \exp(-\varepsilon_C(\cdot))$, where $\varepsilon_C(\cdot)$ is called the energy function of the clique $C$. The smaller the value of $\varepsilon(\boldsymbol{A}_C = \boldsymbol{a}_C)$, the more likely that $\boldsymbol{A}_C = \boldsymbol{a}_C$. $Z$ is the partition function that ensures that $P(\boldsymbol{A})$ follows the probability distribution:

$$Z = \sum_{\boldsymbol{A}} \prod_{C \in \mathcal{C}} \phi_C(\boldsymbol{A}_C) \tag{2}$$

EMKN can be transformed to MRF immediately if its nodes are regarded as random variables. We still use $\boldsymbol{X} = \{X_1, X_2, \cdots, X_m\}$ and $\boldsymbol{Y} = \{Y_1, Y_2, \cdots, Y_n\}$ to denote these variables of $G = (\boldsymbol{X}, \boldsymbol{Y})$. All of $Y_i$ and $X_j$ can take real values from -1 to 1, which indicates the degree of these entities on one patient: we can assign the entities with positive assertion as 1, negative as -1, and others modifiers as 0.5. Entities which do not appeared are set as 0. Inspired by the Ising model, we define the energy function over the nodes $Y_i$ and $X_j$ here as

$$\varepsilon(Y_i = y_i, X_j = x_j) = f(Y_i, X_j) \cdot (y_i \cdot x_j) \tag{3}$$

$f(Y_i, X_j)$ is a function with value independent of the assignment of $Y_i$ and $X_j$. When $f(Y_i, X_j) < 0$, the model prefers $Y_i$ and $X_j$ taking the same sign. Conversely, $f(Y_i, X_j) > 0$ implies that $Y_i$ and $X_j$ are more likely to have different signs. Without ambiguity, we also call $f(Y_i, X_j)$ as an energy function.

### 3.3 Inference on MRF

Given a set of observed variables $\boldsymbol{x} = \{x_1, x_2, \cdots, x_m\}$ of one patient, we calculated the probability of $P(Y_i = 1 | \boldsymbol{X} = \boldsymbol{x})$ and ranked $Y_i \in \boldsymbol{Y}$ accordingly as the result. This is a kind of inference task.

We first calculated $P(\boldsymbol{y}|\boldsymbol{x})$ with the probability distribution defined above.

$$P(\boldsymbol{y}|\boldsymbol{x}) = \frac{P(\boldsymbol{y}, \boldsymbol{x})}{\sum_{\boldsymbol{Y}} P(\boldsymbol{Y}, \boldsymbol{x})} \tag{4}$$

Notice that all the cliques in $G = (\boldsymbol{X}, \boldsymbol{Y})$ are edges, then we get

$$P(\boldsymbol{y}, \boldsymbol{x}) = \frac{1}{Z} \prod_{i,j, <Y_i, X_j> \in E} \phi(y_i, x_j), \tag{5}$$

where $E$ is the set of edges. Plugging Eq. (5) into Eq. (4), we obtain

$$\begin{aligned} P(\boldsymbol{y}|\boldsymbol{x}) &= \prod_i \frac{\prod_{j, <D_i, S_j> \in E} \phi(y_i, x_j)}{\sum_{Y_i} \prod_{j, <Y_i, X_j> \in E} \phi(Y_i, x_j)} \\ &= \prod_i P(y_i|\boldsymbol{x}) \end{aligned} \tag{6}$$

Eq. (6) shows that the value of $Y_i$ are independent of not only any symptoms which are not its neighbor, but also all other diseases $Y_{\setminus i}$. This allow us to calculate the probability of each disease separately.

$$P(y_i|\boldsymbol{x}) = \frac{1}{Z(\boldsymbol{x})}\phi(y_i,\boldsymbol{x}) = \frac{1}{Z(\boldsymbol{x})}\prod_{j=1}^{|\boldsymbol{X}|}\phi(y_i,x_j) = \frac{1}{Z(\boldsymbol{x})}\exp[-\sum_{j=1}^{|\boldsymbol{X}|}\varepsilon(y_i,x_j)] \quad (7)$$

where $Z(\boldsymbol{x}) = \sum_{Y_i}\phi(Y_i,\boldsymbol{x})$ is the partition function.

### 3.4 Parameter Learning on MRF

Once we transformed the EMKN to MRF and determined the appropriate inference method, the last step was to learn the parameters of the potential function from the training data. For clarity, we introduce the learning process here with $f_\theta(Y_i, X_j) = \theta_{ij}$; it can still be an arbitrary differential function.

Similar to many discriminative models, we learn the parameter $\boldsymbol{\theta}$ by maximizing the likelihood function of training data:

$$\begin{aligned}L(\boldsymbol{\theta}) &= \sum_{k=1}^{K}\sum_{i=1}^{|\boldsymbol{Y}|}\ln P(y_i^{(k)}|\boldsymbol{x}^{(k)}) - \sum_{i=1}^{|\boldsymbol{Y}|}\sum_{j=1}^{|\boldsymbol{X}|}\frac{\theta_{ij}^2}{2\sigma^2}\\ &= \sum_{k=1}^{K}\sum_{i=1}^{|\boldsymbol{Y}|}(\varepsilon(y_i^{(k)},\boldsymbol{x}^{(k)}) - \ln Z(\boldsymbol{x})) - \sum_{i=1}^{|\boldsymbol{Y}|}\sum_{j=1}^{|\boldsymbol{X}|}\frac{\theta_{ij}^2}{2\sigma^2}\end{aligned} \quad (8)$$

where $K$ is the number of training records. The second term is a Gaussian prior over the parameters $\boldsymbol{\theta}$. Here, we use the stochastic gradient descend (SGD) to optimize $L(\boldsymbol{\theta})$. The log-likelihood of one single instance is

$$l(\boldsymbol{\theta}) = -\sum_{i=1}^{|\boldsymbol{Y}|}\sum_{j=1}^{|\boldsymbol{X}|}\varepsilon(y_i,x_j) - \sum_{i=1}^{|\boldsymbol{Y}|}\ln Z(\boldsymbol{x}) - \frac{1}{K}\sum_{i=1}^{|\boldsymbol{Y}|}\sum_{j=1}^{|\boldsymbol{X}|}\frac{\theta_{ij}^2}{2\sigma^2} \quad (9)$$

The partial derivation of $l(\boldsymbol{\theta})$ with respect to $\theta_{ij}$ is

$$\frac{\partial}{\partial \theta_{ij}}l(\boldsymbol{\theta}) = -\frac{\partial}{\partial \theta_{ij}}\varepsilon(y_i,x_j) - \frac{1}{Z(\boldsymbol{x})}\frac{\partial}{\partial \theta_{ij}}Z(\boldsymbol{x}) - \lambda\theta_{ij} \quad (10)$$

where $\lambda = -1/K\sigma^2$. For brevity, we let

$$g(y_i, x_j) = -\frac{\partial}{\partial \theta_{ij}} \varepsilon(y_i, x_j) \tag{11}$$

then

$$\begin{aligned}
\frac{\partial}{\partial \theta_{ij}} l(\boldsymbol{\theta}) &= g(y_i, x_j) - \sum_{Y_i} [\frac{\exp(-\sum_j \varepsilon(y_i, x_j))}{Z(X)} \cdot g(y_i, x_j)] - \lambda \theta_{ij} \\
&= g(y_i, x_j) - \sum_{Y_i} [P(y_i|X) \cdot g(y_i, x_j)] - \lambda \theta_{ij} \\
&= g(y_i, x_j) - \boldsymbol{E}_{P(y_i|X)}[g(y_i, x_j)] - \lambda \theta_{ij}
\end{aligned} \tag{12}$$

where $\boldsymbol{E}_P[X]$ is the expectation of $X$ under the distribution $P$. Once we obtain the partial derivative of $l(\boldsymbol{\theta})$, we can update $\theta_{ij}$ with a proper learning rate $\eta$

$$\theta_{ij} \leftarrow \theta_{ij} + \eta \frac{\partial}{\partial \theta_{ij}} l(\theta) \tag{13}$$

It is ostensibly necessary to calculate all the $|\boldsymbol{Y}|$ Y-type entities to determine the likelihood in Eq. (8), which is cumbersome and time-consuming. To accelerate the training speed, we sampled the negative $Y$ with the same number of the positive $Y$ from the top-$k$ neighbor list of the given positive $X$, as measured by the energy function $f(Y_i, X_j)$, to increase the sample possibility of negative $Y$ with high confidence to be positive.

## 4. Distributed Medical Entity Representation

In last section, we represented each medical entity with an individual node, which is not reasonable. For example, "diabetes" is more similar to the "type II diabetes" than "pneumonia"-this similarity should be reflected in the entity representation. KRL methods are designed to capture this similarity to some degree, by embedding the entities to a low-dimensional, dense vector space.

## 4.1 Knowledge representation learning

The general idea of KRL is also to construct an energy function for triples. A valid triple has lower energy while an invalid triple has higher energy. The representations of entities and relationships are tuned for this purpose. The LFM model computes the energy of a triple $(e_h, r, e_t)$ by

$$g_{\text{LFM}}(e_h, r, e_t) = -\mathbf{e_h}^T \mathbf{W_r} \mathbf{e_t} \tag{14}$$

where $\mathbf{W_r} \in \mathbb{R}^{d \times d}$ is a transformation matrix and $\mathbf{e_h}, \mathbf{e_t} \in \mathbb{R}^d$ are embeddings of the entities.

The TransE model treats relationships as translations between two entities. For a valid triple $(e_h, r, e_t)$, it hopes that the embeddings satisfy $\mathbf{e_h} + \mathbf{r} \approx \mathbf{e_t}$. The energy function is the distance of two vectors:

$$g_{\text{TransE}}(e_h, r, e_t) = |\mathbf{e_h} + \mathbf{r} - \mathbf{e_t}|_{L_1/L_2} \tag{15}$$

## 4.2 Medical knowledge representation

Inspired by the above two knowledge representation models, we introduced the distributed representation of medical concepts to the MRF inference architecture.

Here, we use $\mathbf{y_i}, \mathbf{x_j} \in \mathbb{R}^d$ to denote the embeddings of $Y_i$ and $X_j$, and define the LFM and the Trans model over this pair as

$$f_{\text{LFM}}(Y_i, X_j) = -\text{norm}(\mathbf{y_i}^T \mathbf{W_{xy}}) \mathbf{x_j} \tag{16}$$

$$f_{\text{Trans}}(Y_i, X_j) == |\mathbf{y_i} + \mathbf{r_{xy}} - \mathbf{x_j}|_{L_2} + \gamma \tag{17}$$

$\text{norm}(\mathbf{x}) = \mathbf{x}/|\mathbf{x}|$, $\mathbf{W_{xy}} \in \mathbb{R}^{d \times d}$ is the transformation matrix, and $\mathbf{r_{xy}} \in \mathbb{R}^d$ is the relation embedding. $\gamma$ is a constant bias which ensures that the $f_{\text{Trans}}$ can be negative. We arbitrarily set $d = 100$ and $\gamma = -1$. To alleviate overfitting, we constrained the norm of each entity embedding as 1 after each update.

The parameter learning process is similar to that described in Section 3.4 once we obtained $g(y_i, x_j)$ according to Eq. (11). This learning process is more difficult, however, because the energy functions are coupled together. Once one entity embedding is updated, all the energy values related to this entity change. In Section 3.4, where $f_\theta(y_i, x_j) = \theta_{ij}$, other energy values remain unchanged despite of the update of $\theta_{ij}$.

## 5. Experiments and evaluation

### 5.1 Experiments Setup

We randomly selected 700 records as a training set, which was used for EMKN re-construction and parameter learning, and reserved the remaining 292 records as a test set. Experiments were run to evaluate the inference and learning capacity of the MRF-based EMKN on three tasks. The corresponding subgraph and training and test data statistics for each task are listed in Table 2. For each test record, we took the $\boldsymbol{X}$ in subgraph as an input to predict the possibility of $Y_i = 1$ for each $Y_i \in \boldsymbol{Y}$, then re-ranked $\boldsymbol{Y}$ accordingly as result. The golden-standard assigned entities in $\boldsymbol{Y}$ with non-negative modifiers as 1, and others as 0.

Table 2: Training and test data statistics for three tasks.

| task | subgraph | training data | | | test data | | |
|---|---|---|---|---|---|---|---|
| | | data size | x per record | y per record | data size | x per record | y per record |
| symptom/testresult→disease | SD-EMKN | 660 | 6.49 | 3.79 | 186 | 4.9 | 3.05 |
| disease→treatment | DTr-EMKN | 509 | 4.03 | 6.89 | 161 | 3.36 | 6.89 |
| symptom→test | ST-EMKN | 594 | 3.96 | 4.88 | 165 | 3.55 | 3.91 |

We discarded the training and test records without any positive entity in $\boldsymbol{X}$ or $\boldsymbol{Y}$. New entities also occasionally appeared during the test process, but the knowledge needed is beyond the scope of our EMKN, so we also discarded any test instances with more than half of the new X-type entities.

We used the three energy functions listed above for comparison. In function $f_{\text{LFM}}$ and $f_{\text{Trans}}$, we adopted different representations for the same entity in different tasks to explore the potential ability of models, although the existence of **W** and **r** allow them to keep consistent. We also designed another three lazy functions as a baseline, which means that they did not have parameters to be learned:

$$\begin{cases} f_{\text{weight}}(Y_i, X_j) = w_{ij} \\ f_{\text{log-w}}(Y_i, X_j) = \log_2(w_{ij} + 1) \\ f_{\text{TF-IDF}}(Y_i, X_j) = \log_2(w_{ij} + 1) \times \log \frac{|\boldsymbol{X}|}{\deg(Y_i)} \end{cases} \tag{18}$$

where $w_{ij}$ is the weight between $Y_i$ and $X_j$. If the edge does not exist, $w_{ij} = 0$. $\deg(Y_i)$ is the degree of $Y_i$. $f_{\text{log-w}}$ add a log-linear penalty for $f_{\text{weight}}$ and $f_{\text{TF-IDF}}$ consider the degree of $Y_i$ further, inspired by TF-IDF: The more neighbors a $Y$ entity has, the weaker that its relationship with each neighbor.

The baseline methods above still use the MRF inference architecture, so we implemented another three baseline models using naïve Bayes, neural networks, and logistic regression. For each record, we regarded $\boldsymbol{X}$ as a feature set, and represented it using a sparse vector. Then we trained individual binary classifier for each $Y \in \boldsymbol{Y}$. Utilizing these methods directly in this way yields very poor results due to the high feature dimension and the class imbalance, so we applied two pre-processing steps. We first sampled the same number of negative instances as the positive instances for each $Y$. The negative instances which had positive features overlapping with the positive instances were preferred. We then removed the features that were 0 for all the selected training instances to reduce the feature dimension.

### 5.2 Evaluation measures

Diagnostic support systems for specific diseases have many standard evaluation measures, like ROC curve or AUC, which can not be applied to our evaluations directly. Returning the most probable medical entities from a fixed entity set is more akin to an information retrieval (IR) task. We instead used P@k, R@k, and average precision (AP) to evaluate the performance on single test instance.

P@k defines the fraction of true positive Y entities

$$P@k = \frac{\#(\text{true positive Ys returned in top-k items})}{k} \qquad (19)$$

This measure is meaningless for an arbitrary $k(k = 10$, for example). When there is only one positive Y in the test record, its P@10 can no longer more than 0.1. Therefore, we would assign $k$ as the exact number of positive diseases in the evaluated test record.

R@k defines the fraction of relevant $Y$s that are returned in the top-k items

$$R@k = \frac{\#(\text{true positive Ys returned in top-k items})}{\#(\text{positive diseases in records})} \qquad (20)$$

R@k is not a standard evaluation measure in IR because the denominator, which was easy to obtain in our experiment, is hard to estimate in the real retrieval pool. We set $k = 10$ during the evaluation.

AP is the average precision value at the entity list after each true positive entity is returned. That is, if the number of positive $Y$ is $m$, and we return $n$ of them, ranked as $r_1, r_2, \cdots, r_n$, respectively, then the $AP$ is given by

$$AP = \frac{1}{m} \sum_{i=1}^{n} \frac{i}{r_i} \qquad (21)$$

The most ideal condition is that the $m$ results are all returned and ranked at the top of the list, then $AP = 1$.

We also used Mean P@k(MP@k), mean R@k(MR@k), and mean average precision(MAP) measures to evaluate over the whole test set. They are the mean values of the three above measures among all the test instances $\boldsymbol{Q}$:

$$\text{MP@k}(\boldsymbol{Q}) = \frac{1}{|\boldsymbol{Q}|} \sum_{j=1}^{|\boldsymbol{Q}|} P@k(Q_j)$$

$$\text{MR@k}(\boldsymbol{Q}) = \frac{1}{|\boldsymbol{Q}|} \sum_{j=1}^{|\boldsymbol{Q}|} R@k(Q_j)$$

$$\text{MAP}(\boldsymbol{Q}) = \frac{1}{|\boldsymbol{Q}|} \sum_{j=1}^{|\boldsymbol{Q}|} AP(Q_j)$$

## 6. Results and Discussion

### 6.1 Evaluation results

The mean evaluation measures of different methods and energy functions for three tasks are listed in Table 3. We prefer to use the R@10 as the primary evaluation measure, so the percentages of test instances with R@10 above 0.1 and 0.9 are also listed in this table. The distribution of these measures over the whole test set is shown in Figure 2.

In initial diagnosis task, there were 80.11% of the test records with R@10 above 0.1, indicating that these records returned at least 1 records in the top-10 results. In test and treatment recommendation tasks, this percentage was 87.88% and 92.55%.

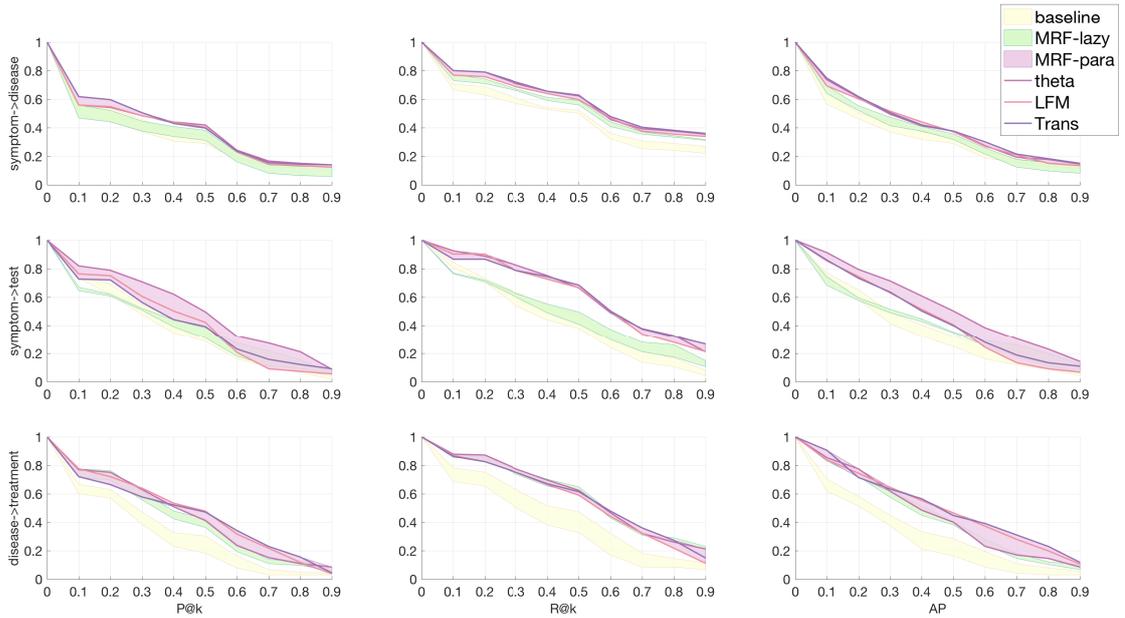

Figure 2: The distribution of measures. Three column show the distribution of P@k, R@k and AP, respectively. Three rows represent the three inference tasks. The x-axis of each subfigure represents the values of each measure, and the y-axis value is the cumulative percentage of records which achieves the performance higher than x-axis value. For clarity, we only draw the results distribution of last three methods in Table 3. The color blocks fill the upper bound and the lower bound of three groups of results: the baseline methods, and the MRF energy functions with or without parameters.

Table 3: Evaluation measures of different methods and energy functions for three tasks.

| Methods | Symptom/TestResult→ Disease | | | | |
| --- | --- | --- | --- | --- | --- |
| | MP@R | MAP | MR@10 | R@10>0.1 | R@10>0.9 |
| Naïve Bayes | 0.3314 | 0.3457 | 0.4707 | 0.7043 | 0.2688 |
| Logistic | 0.3011 | 0.3186 | 0.4468 | 0.6935 | 0.2204 |
| Neural network | 0.2979 | 0.3122 | 0.4537 | 0.6667 | 0.2527 |
| Weight | 0.2532 | 0.3191 | 0.5096 | 0.7312 | 0.3118 |
| Log-weight | 0.2956 | 0.351 | 0.5239 | 0.7634 | 0.3172 |
| TF-IDF | 0.3224 | 0.369 | 0.5321 | 0.7742 | 0.3172 |
| Theta | 0.3431 | 0.391 | 0.5658 | **0.8011** | 0.3548 |
| LFM | 0.3359 | 0.383 | 0.5472 | 0.7688 | 0.3387 |
| Trans | **0.3543** | **0.4044** | **0.5728** | **0.8011** | **0.3602** |

| Methods | Symptom→ Test | | | | |
| --- | --- | --- | --- | --- | --- |
| | MP@R | MAP | MR@10 | R@10>0.1 | R@10>0.9 |
| Naïve Bayes | 0.3537 | 0.3598 | 0.4285 | 0.8075 | 0.0745 |
| Logistic | 0.3322 | 0.3212 | 0.3893 | 0.8447 | 0.0435 |
| Neural network | 0.3813 | 0.3709 | 0.4439 | 0.8261 | 0.0807 |
| Weight | 0.3234 | 0.3659 | 0.4164 | 0.764 | 0.1056 |
| Log-weight | 0.3438 | 0.3816 | 0.4395 | 0.7702 | 0.1304 |
| TF-IDF | 0.3641 | 0.4002 | 0.4606 | 0.764 | 0.1491 |
| Theta | **0.4701** | **0.5123** | **0.5939** | **0.9255** | 0.2112 |
| LFM | 0.3755 | 0.4175 | 0.5769 | 0.903 | 0.2121 |
| Trans | 0.3754 | 0.4402 | 0.5894 | 0.8667 | **0.2667** |

| Methods | Disease→ Treatment | | | | |
| --- | --- | --- | --- | --- | --- |
| | MP@R | MAP | MR@10 | R@10>0.1 | R@10>0.9 |
| Naïve Bayes | 0.3057 | 0.3105 | 0.4314 | 0.7576 | 0.097 |
| Logistic | 0.2985 | 0.2979 | 0.4189 | 0.7818 | 0.1152 |
| Neural network | 0.2608 | 0.2584 | 0.3416 | 0.6848 | 0.0667 |
| Weight | 0.3567 | 0.3979 | 0.5455 | 0.8545 | 0.2242 |
| Log-weight | 0.3524 | 0.4008 | 0.5422 | 0.8545 | **0.2303** |
| TF-IDF | 0.3969 | 0.4288 | **0.5573** | 0.8727 | 0.2121 |
| Theta | 0.3971 | 0.4258 | 0.5568 | **0.8788** | 0.2121 |
| LFM | **0.4161** | 0.4669 | 0.5189 | 0.8696 | 0.1118 |
| Trans | 0.408 | **0.4822** | 0.5414 | 0.8634 | 0.1491 |

Generally speaking, the MRF baseline methods outperformed other machine-learning baseline methods. The performance of these baseline methods was enhanced as the energy function complexity increased. After adding parameters to the energy functions, the performance was even further enhanced.

For the diagnostic support system, the Trans model showed optimal performance in all three measures. For the test recommendation task, learning the $\boldsymbol{\theta}$ directly from the data was optimal. For the treatment plan recommendation task, the performance of different energy functions varied across different evaluation measures.

### 6.2 Medical entity visualization

During the training process of $f_{\text{LFM}}$ and $f_{\text{Trans}}$, we obtained the distributed representation of medical entities. These medical embeddings were expected to capture the similarity among entities in regards to knowledge level. To verify this, we reduced the dimension of embeddings from 100 to 2 using the t-Distributed Stochastic Neighbor Embedding (t-SNE) technique for visualization. Figure 3 shows the disease embeddings obtained by Trans model on the diagnosis task. The disease entities are colored in the figure according to the first letter of their ICD-10 code, which indicates their corresponding ICD section. The complete ICD-codes of several disease entities are provided in the enlarged block. It can be seen that similar medical entities indeed stayed close together in the vector space.

We also list several symptom-disease pairs in Figure 4 to illustrate the translation relationships of these entities in vector space. As expected, these relationships are tried to keep parallel to each other.

### 6.3 Discussion

The three MRF baseline methods use lazy mechanisms and are not equipped with any explicit parameters or learning process. We only need to construct the EMKN from the medical records set, and then calculate the energy values based on the graph measures. These functions are suitable for the online learning of massive flows of data, e.g., for updating the EMKN using daily medical records or other abundant sources of medical knowledge. The performance of these energy functions increased gradually over the course of our experiment, demonstrating the effectiveness of our method in regards to improving them.

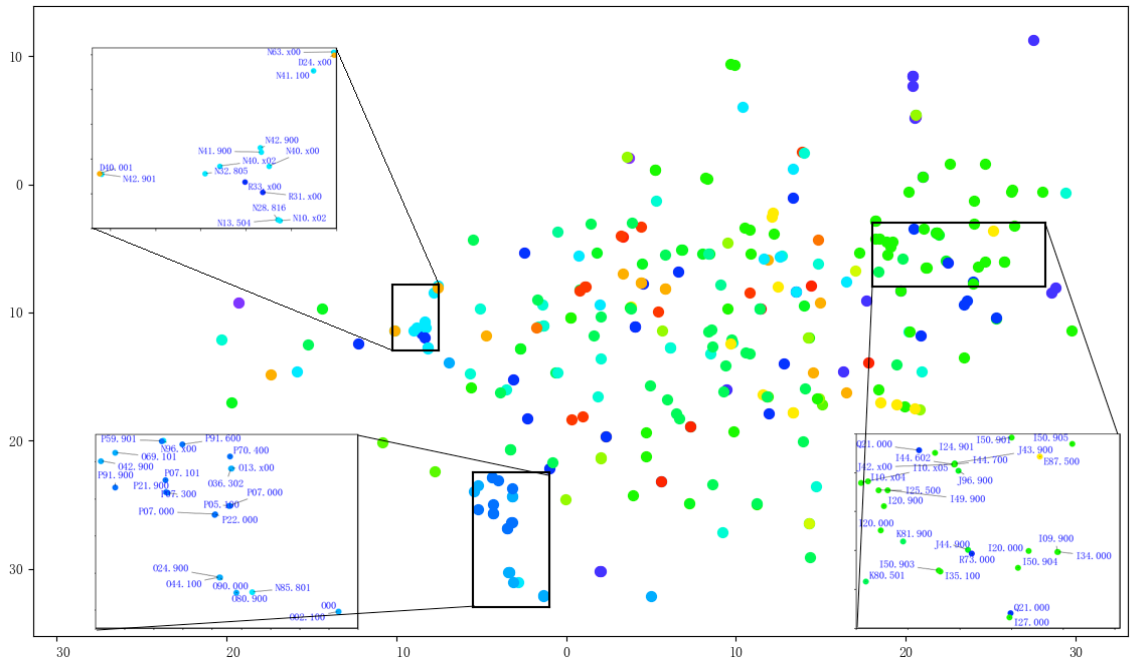

Figure 3: The visualization of disease entities of Trans model.

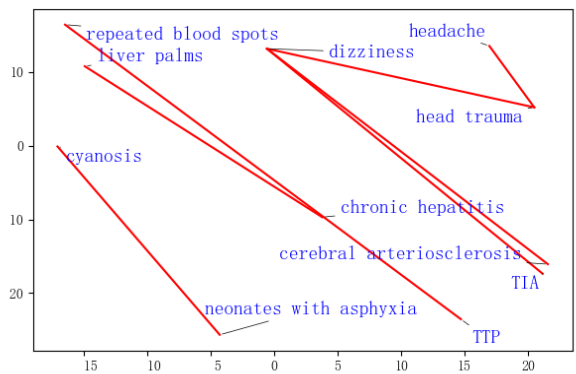

Figure 4: Translation relationships of symptom-disease pairs obtained via Trans model.

Conversely, the other three MRF energy functions we tested involving parameters, which were learned iteratively from training data. They also performed better than lazy learners. No energy function, however, outperformed others on all three tasks. This differences in performance of the same energy function in different tasks indicates that the energy functions should be designed individually for each task, because each function has its own unique characteristics and application scenarios.

The $f_\theta$ sets parameters on the edges of the EMKN, where each edge $<Y_i, X_j>$ has one parameter $\theta_{ij}$. An advantage is that each energy function is independent of the others. The update of $\theta_{ij}$ only affects the value of $f(Y_i, X_j)$. However, its parameter size is $O(n^2)$, where $n$ is the number of nodes - this is fairly large compared to our data set.

$f_{\text{LFM}}$ and $f_{\text{Trans}}$ move the parameters from edges to nodes, by learning the distributed representation of medical entities. This reduces the parameter size to $O(n)$. Smaller parameter size is helpful to reduce the model complexity and alleviate overfitting. This also ensure that entities are no longer independent of each other. Similar entities are close in the vector space, as shown above. Additionally, the representation of entities with low occurrence frequency is affected by the high-frequency entities, which can improve the generalization power of the model on low-frequency knowledge.

The one-to-one relationship assumption is a persistent problem. That is, it is assumed that one head entity is related exactly to one tail entity. These models perform poorly when the real relationships of entities are far from satisfied the assumption. For example, the blood pressure is a common test item for many symptoms with a degree in ST-EMKN is up to 301. As shown in Table 1, the degree median of test entities in ST-EMKN is smaller than that of the diseases in SD-EMKN, but the degree-mean is larger - in effect, there are several test items having extremely large degree. It is hard to learn the embeddings for these test items' symptom neighbors to satisfy the assumed relationships unless they overlap. For $f_\theta$, however, we only need to increase the weight parameter between the blood pressure and its corresponding symptoms, while other energy functions are not be affected. Therefore, the $f_\theta$ performed best on test recommendation task. Another unsatisfactory example of diagnosis task performance is shown in Figure 4. Both head trauma and cerebral arteriosclerosis can cause dizziness, but their embeddings should not be close whatsoever. A better way to alleviate the dissatisfaction of Trans assumption is to subdivide the symptom-disease relationships further, maybe according to the department or the ICD section. Trans and LFM are, after all, multi-relational models.

There is one more limitation to EMKN representation worth noting. The edges between entities represent co-occurrence relationships rather than cause and effect relations, which renders many edges redundant or unnecessary. The existence of these edges makes the model more prone to overfitting.

## 7. Conclusion

This work is a preliminary attempt to establish general CDSS. We developed a new EMR-driven medical knowledge representation and inference system, with the EMKN, MRF, and representation learning techniques. We used the the current condition of one patient as an input to obtain corresponding recommendations for medical tests, possible diseases, and treatment plans. Six energy functions were proposed and actual clinical records were utilized to evaluate the performance.

The MRF-based inference module outperformed other machine learning baseline methods. The performance was further improved after we introduced the parameters to energy functions. The best system in the diagnosis support task guaranteed that 80.11% of the test records returned at least one right disease out of the top-10 results; these percentages were 87.88% and 92.55% for test and treatment recommendations, respectively. The medical entity embeddings were obtained and evaluated for the expected similarity in knowledge level. None of the methods we tested outperformed all other methods on all tasks, however, suggesting that the energy function should be individually designed for each task.

In the future, we plan to further refine the entity relationships and the energy functions.

## Acknowledgment

We thank the Second Affiliated Hospital of Harbin Medical University for providing the corpus used in this study.